\documentclass[a4paper,fleqn]{cas-dc}

\usepackage[authoryear,longnamesfirst]{natbib}

\usepackage{algpseudocode}
\usepackage{algorithm}

\usepackage{tikz}
\usepackage{xcolor}
\usepackage{pgfplots}
\definecolor{bblue}{HTML}{4F81BD}
\definecolor{rred}{HTML}{C0504D}
\definecolor{ggreen}{HTML}{9BBB59}
\definecolor{ppurple}{HTML}{9F4C7C}
\definecolor{oorange}{HTML}{EC700A}
\definecolor{tteal}{HTML}{0A8C98}

\def\tsc#1{\csdef{#1}{\textsc{\lowercase{#1}}\xspace}}
\tsc{WGM}
\tsc{QE}
\tsc{EP}
\tsc{PMS}
\tsc{BEC}
\tsc{DE}
\begin{document}
\let\WriteBookmarks\relax
\def\floatpagepagefraction{1}
\def\textpagefraction{.001}

\shorttitle{Online Grooming Context Determination}

\shortauthors{Street et~al.}

\title [mode = title]{Enhanced Online Grooming Detection Employing Context Determination and Message-Level Analysis}                      


\author[1]{Jake Street}[orcid=0000-0001-7511-2910]

\ead{jake.street02@ntu.ac.uk}

\cormark[1]


%\ead{cvr_1@tug.org.in}

%\ead[url]{www.cvr.cc, cvr@sayahna.org}

\credit{Conceptualization of this study, Methodology, Investigation, Writing - Original Draft}

\affiliation[1]{organization={Nottingham Trent University},
    addressline={50 Shakespeare St}, 
    city={Nottingham},
    % citysep={}, % Uncomment if no comma needed between city and postcode
    postcode={NG1 4FQ}, 
    country={United Kingdom}}

\author[1]{Isibor Kennedy Ihianle}[orcid=0000-0001-7445-8573]
\credit{Supervision, Writing - Review \& Editing}

\ead{isibor.ihianle@ntu.ac.uk}

\author[1]{Funminiyi Olajide}[orcid=0000-0003-1627-6637]
\credit{Supervision, Writing - Review \& Editing}

\ead{funminiyi.olajide@ntu.ac.uk}

\author[1]{Ahmad Lotfi}[orcid=0000-0002-5139-6565]
\credit{Supervision, Writing - Review \& Editing}
\ead{ahmad.lotfi@ntu.ac.uk}
\cortext[cor1]{Corresponding author}
% \cortext[cor2]{Principal corresponding author}

% \fntext[fn1]{This is the first author footnote. but is common to third
% \fntext[fn2]{Another author footnote, this is a very long footnote and
%   it should be a really long footnote. But this footnote is not yet
%   sufficiently long enough to make two lines of footnote text.}

% \nonumnote{This note has no numbers. In this work we demonstrate $a_b$
%   the formation Y\_1 of a new type of polariton on the interface
%   between a cuprous oxide slab and a polystyrene micro-sphere placed


\begin{abstract}
Online Grooming (OG) is a prevalent threat facing predominately children online, with groomers using deceptive methods to prey on the vulnerability of children on social media/messaging platforms. These attacks can have severe psychological and physical impacts, including a tendency towards revictimization. Current technical measures are inadequate, especially with the advent of end-to-end encryption which hampers message monitoring. Existing solutions focus on the signature analysis of child abuse media, which does not effectively address real-time OG detection. This paper proposes that OG attacks are complex, requiring the identification of specific communication patterns between adults and children. It introduces a novel approach leveraging advanced models such as BERT and RoBERTa for Message-Level Analysis and a Context Determination approach for classifying actor interactions, including the introduction of Actor Significance Thresholds and Message Significance Thresholds. The proposed method aims to enhance accuracy and robustness in detecting OG by considering the dynamic and multi-faceted nature of these attacks. Cross-dataset experiments evaluate the robustness and versatility of our approach. This paper's contributions include improved detection methodologies and the potential for application in various scenarios, addressing gaps in current literature and practices.

%OG is a prevalent threat facing predominately children online, with groomers using deceptive methods to prey on the vulnerability of children on social media/messaging platforms. This investigation proposes the use of a novel approach - context determination, to detect OG across the 'perverted-justice.com' (PJ) and PAN12 datasets. The Message Level Analysis results showed that models tended to perform well when taking an inter-dataset approach (88.8\%), however these models tended to lack robustness when taking a cross-dataset approach (67.2\%). However, this is able to be overcome by taking a context determination approach giving the same dataset F1 score of 0.973 and a different dataset F1 score of 0.960 when using RoBERTa with a Message Threshold of 0.45 and Actor Significance Threshold (AST) of 0.05. 
% It is proposed that this approach allows SMPs to easily tune values to obtain a False Positive/False Negative value that fits with their use case as well as combining context determination with other insights to provide further details, so less intervention is needed from human reviewers.
% Above commented out due to 150 word limit on abstract

\end{abstract}

\begin{highlights}
\item Transformer-based models are most effective for Adult/Child determination at the message-level within an Online Grooming Context
\item Context determination improves robustness of Online Grooming detection from using a cross-dataset approach
\item Context Determination allows for the adjustment of False Positive/Negative frequencies to meet differing Human Reviewer/Social Media platform use cases
\end{highlights}

% Keywords

\maketitle{}

\section{Introduction}

Online Grooming (OG) attacks are well known throughout the public and within psychology-focused literature, with children being taught online safety skills to mitigate these attacks as part of the national curriculum \citep{dfe2023}. Despite the awareness of these attacks, they are still expected to be prevalent with 29\% of children meeting someone they have met online who they do not know in real life \citep{lobe2011}. 
The effectiveness of education as a mitigation method is not known however these attacks tend to have a significant psychological and physical impact on the child as detailed by \citet{gov2021}. This also includes an affinity to `revictimisation', in which people who have experienced abuse are likely to behave in a way that victimises themselves further \citep{butler2020}. Throughout literature there have been many effective attempts at identifying OG attacks in a binary fashion \citep{villatorotello2012,milonflores_2022_how,borj_2023_detecting}. However, social media platforms (SMPs) fail to implement satisfactory solutions to this problem with technical measures. In addition, the introduction of features such as end-to-end encryption on these platforms \citep{homeoffice2023} any potential mitigation method is even more difficult to implement, due to the inability of the platform to access messages/transcripts. Because of this, it is suggested that an endpoint solution would be required. However, as determined by \citet{street_2023_evaluating}, an endpoint text-analysis solution using Optical Character Recognition was found to be ineffective at detecting OG attacks on frequently used SMPs such as Twitter and Facebook Messenger, exacerbating the challenge of addressing these attacks. Solutions implemented to tackle the effects of OG typically focus on the signature analysis of child abuse media on SMPs \citep{homeoffice2023}. However, this approach does not address OG attacks in real-time, and it is uncertain if end-to-end encryption will render this method ineffective. Therefore, it is essential to consider the additional complexities surrounding OG and develop solutions to overcome these limitations. 
It is proposed that OG attacks are more complex than other text-based social engineering attacks, as determining an OG attack from a single message is difficult due to the need to meet multiple criteria, such as establishing that an adult and a child are communicating. This complexity is further compounded by the differing goals of attackers. According to \citet{chiu2018}, there are two main types of attackers: `Fantasy Child Sex Offenders' (FCSOs), who focus on online contact and immediate sexual gratification, and `Child Contact Sex Offenders' (CCSOs), who aim to arrange a physical meet-up with the victim. When referring to OG attacks within literature these are predominately agreed to be text-based attacks between an adult and a child for the adult's sexual gratification. Research also suggests that emotional gratification may be a motive for some attackers \citep{broome2020}.

To address this, this paper aims to determine whether an adult and a child are communicating within a transcript. This topic has been explored in the context of author profiling determination, which is common in literature \citep{goswami2009,nguyen2021,peersman2011,argamon2009}. However, many methodologies use blogging websites for author profiling due to their vast datasets and labeled data, rather than focusing on the context of OG (OG). While author profiling has been somewhat considered in the context of OG within the PAN13 competition \citep{rangel2013}, the robustness of these approaches needs further application to OG detection.
Robustness is a key consideration that needs to be made towards any OG solution. In the context of this paper, there are limitations with datasets currently available due to the language used in these transcripts likely being outdated in terms of their `textspeak' as well as the topics of conversation that are discussed.
In addition, there are likely changes in the method of initiation that is used by the attacker due to the decrease in usage of `chatroom-based' SMPs and the increase in `reel-based' SMPs such as TikTok and Instagram \citep{ofcom2023}. Current messaging on these new-age SMPs contain features/data that were not previously available; replies to a specific message, the sending of media in conversation \& the ability to reply to media in chat, destructive messages/media, and video calls. Whereas focusing on text-based messaging alone is likely to have real world application, it is possible that the presence additional features within a given transcript provides a greater deal of information that will allow for a greater accuracy of determinations.
To date, no method has utilized identifying the Actor-Actor dynamic to classify OG, nor has there been an approach using Actor Significance as a concept to determine OG. 
In addition, the polarisation of message determinations has not been addressed within literature. This paper shall describe this Actor-Actor context determination approach, with the specific contributions of this study being: 

\begin{enumerate}
    \item A novel Message-Level Analysis leveraging advanced models such as BERT and RoBERTa specifically in the context of OG detection.
    \item A novel Context Determination approach for classifying actors and their interactions, which has the potential to allow generalisation to group chat settings.
    %\item Experiments with actor context labelling not previously used within literature.
    \item Introduction of Actor Significance Thresholds and Message Significance Thresholds to better address the specific challenges of OG detection.
    \item Cross-dataset experiments to evaluate the robustness of the proposed context determination approaches and message-level analysis. Furthermore, exploring its potential applicability in different scenarios, demonstrating its versatility and robustness for OG detection.
\end{enumerate}

This paper is organised into several sections. Section 2, an overview of related work will be presented, providing a comprehensive understanding of the current technical and psychological perspectives on OG attacks. Section 3 - the Proposed Methodology section will then offer a theoretical framework for the investigation. This will be followed by the Experimental Methodology Section 4, detailing the configurations and processing methods used, Section 5 presents the results and the discussion allowing for a thorough analysis of the findings. Finally, the Conclusion summarises the outcomes and suggests directions for future research in Section 6.

\section{Related Work}
Initial research into OG attacks primarily focused on identifying the linguistics and psychology of groomers to develop frameworks for understanding these attacks. One of the most prominent frameworks was proposed by \citet{oconnell2003}, which theorized distinct 'phases' of OG, with the severe sexual phases typically occurring towards the end of the communication.
This was validated by \citet{kontostathis2010}, who used K-means clustering on chat logs from Perverted Justice (PJ) \footnote{perverted-justice.com} a vigilante justice website that publishes transcripts of honeypot interactions with offenders. This approach supported the rule-based method known as `chatcoder', categorizing transcript characteristics into eight categories. However, rule-based approaches are likely to experience a high false positive rate when analysing transcripts that resemble but are not indicative of OG, especially when these transcripts are included in datasets containing both negative and positive instances of OG.
Due to the lack of negative OG transcripts within the PJ dataset, it is difficult to validate the real-world application of any approach claiming accuracy based solely on PJ data. Therefore, the PAN12 dataset \citep{inches2012} was employed to broaden the scope and enhance the classification challenge of OG. Most approaches treat this as a typical binary classification problem, employing various text preprocessing methods and training models with different configurations to optimise performance, typically evaluated using F-score metrics across diverse use cases; Identifying Individual Predatory Messages using $\beta = 3$ \citep{inches2012}, and Identifying Predators using $\beta = 0.5$ \citep{inches2012}. Regarding datasets used in the literature, the focus has primarily been on two main datasets. While other studies have explored OG detection methods in different contexts \citep{cheong_2015_detecting,chiang_2019_deceptive,kloess_2015_a}, these investigations do not specifically evaluate the robustness of a potentially generalised OG solution.

Various preprocessing and encoding methods are employed within OG classification tasks, including Word2Vec \citep{ebrahimi_2016_detecting,borj_2023_detecting}, Bag of Words \citep{villatorotello2012,ebrahimi_2016_detecting}, and Term Frequency Inverse Document Frequency (TF-IDF) \citep{villatorotello2012,milonflores_2022_how}. These methods are integral to optimising text data for effective analysis and classification.
In the context of OG detection, preprocessing methods such as stopword removal \citep{cheong_2015_detecting} and repeated fixed letter removal \citep{cheong_2015_detecting} have been utilised. However, the efficacy of these methods has been questioned, particularly regarding common NLP preprocessing techniques like punctuation removal, which may inadvertently remove crucial information used in OG classification \citep{villatorotello2012}. For classification tasks on the PAN12 dataset, a range of models have been explored, including traditional approaches such as Support Vector Machines (SVM) \citep{inches2012}, Naive Bayes (NB) \citep{inches2012,michalopoulos_2011_utilizing}, Logistic Regression \citep{inches2012,rangel2013}, Neural Networks \citep{villatorotello2012,ebrahimi_2016_detecting}, and transformer-based models like BERT/RoBERTa \citep{borj_2023_detecting}. However, research in this area is still in its early stages, with ongoing exploration of model performance and optimisation.
Recent literature on the PJ dataset has primarily focused on clustering methods \citep{chiu2018,kontostathis2010}, owing to its `all positive' nature where all interactions are assumed to involve grooming. These clustering approaches have identified distinct phases within transcripts, aligning with frameworks proposed by previous studies \citep{oconnell2003,edwards2009}.

\section{Methodology}
The critical factor in differentiating OG interactions from other types of interactions lies in reliably determining the context of the interaction. Specifically, this involves establishing whether an adult is interacting with a child and if this interaction is conducted in a grooming manner. Once this context is established, additional indicators such as the use of sexual language can further identify an interaction as OG. Although user profile attributes could provide this information, this method is vulnerable as both attackers and children often falsify their ages on these platforms.

This paper follows two primary methodological steps - a Message-Level Analysis (MLA) for which each phrase within a transcript is scored using a variety of models, including BERT, RoBERTa, SVM, and Naive Bayes (NB); Context Determination - Following the MLA, a comprehensive analysis of each transcript is performed to determine the Full Transcript Context. This process involves evaluating the context at the transcript level for each actor to establish whether the interaction involves both an adult and a child. If such a determination can be made, the transcript is labelled accordingly. Transcripts that clearly involve both an adult actor and a child actor are labelled as OG interactions. The accuracy of this labelling is then analysed. It is important to note that while this approach may be valid within the tested datasets, it may not directly transfer to real-world contexts without additional corroborative insights.

The research questions intended to be answered from this are listed below.

\begin{enumerate}
    \item Can a robust model be formed from taking a full transcript context determination approach? 
    \item What impact do message determination thresholds ($t$ values) and AST values have on the transcript OG determination accuracy?
    \item What model is recommended to be used in different use cases?
\end{enumerate}

% Why each model was selected... (comparison between roberta/bert and NB/svm)

The key piece of information that would help to differentiate an interaction as OG as opposed to other interactions is being able to reliably determine the `context' of the interaction i.e. Is an adult interacting with a child, and is this interaction in a `grooming way'?
Once this piece of information is established the presence of other indicators, e.g. Sexual Language, can help to establish a given interaction as OG. This information could be obtained through attributes on the user's profile however this approach is likely to be easily mitigated with knowledge of this method being made public as well as both attackers and children commonly lying about their age on these platforms.

\begin{figure*}
    \centering
    \includegraphics[width=1\linewidth]{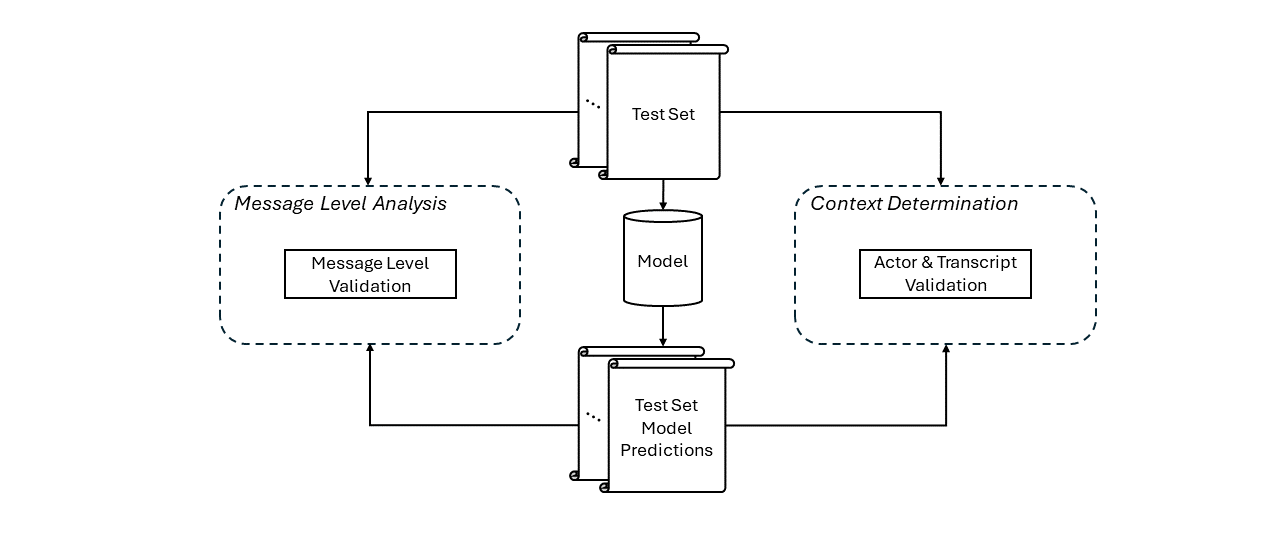}
    \caption{Methodology context.}
    \label{fig:context}
\end{figure*}

\subsection{Message Level Analysis}
\begin{figure*}
    \centering
    \includegraphics[width=1\linewidth]{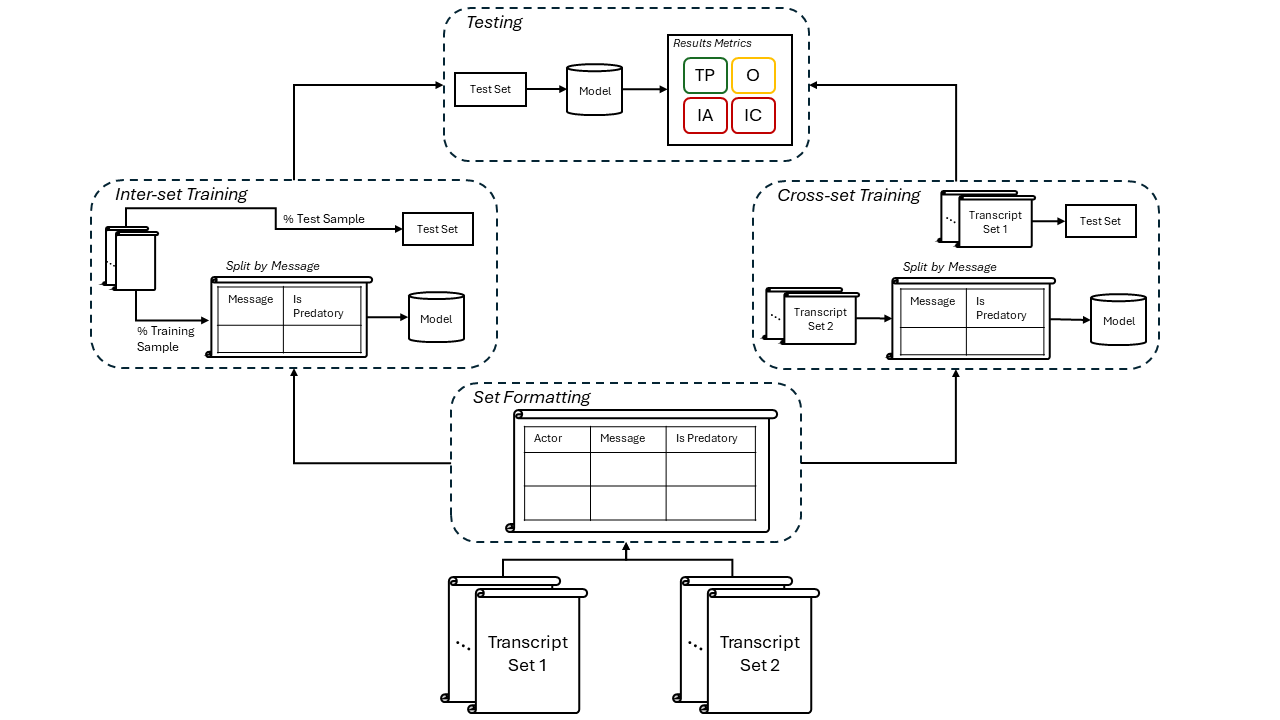}
    \caption{Message Level Analysis (MLA) process showing Inter-Set and Cross-Set approach.}
    \label{fig:MLAmethodology}
\end{figure*}

The Message Level Analysis (MLA) process involves two main approaches as shown within Figure \ref{fig:MLAmethodology} `Inter-set Training', in which testing and training occur on the same set of texts which indicates the accuracy of OG detection within a well defined OG `use case' and `Cross-set Training', in which testing and training occur on two different sets of text which allows observation to be made as to the robustness of a system across different OG use cases.
In the context of a real-world implementation the inter-set experiments provide some indication as to how this would perform on the chatroom attack vector and the cross-set experiments provide some insight into how this will perform at detecting current-day attacks on prevalent SMPs.

As depicted in the `Results Metrics' section of Figure \ref{fig:MLAmethodology}, four metrics are proposed to evaluate the performance of each tested approach. These metrics include True Positive (TP), representing correct `A' or `C' determinations; Incorrect Child Determination (IC), indicating falsely identified `child messages'; Incorrect Adult Determination (IA), denoting falsely identified `adult messages'; and Omissions (O), which counts messages excluded from the test set due to insufficient polarization towards either adult or child determination based on the Message Threshold ($t$) value used. The MLA process focuses solely on classifying messages as either `Adult' or `Child'. It is crucial to consider the types of texts included in this classification, as including non-OG texts may skew message determination values, potentially increasing the False Positive rate in the Context Determination process and minimising the impact of Omissions on the dataset. This assumption requires empirical validation. Therefore, this paper will exclusively analyse OG-positive transcripts within the MLA process, acknowledging the inability to determine whether messages in negative OG transcripts from PAN12 are classified as `Adult' or `Child'.

\subsection{Context Determination}
Context determination aims to identify if an adult and a child are communicating with one another within a transcript. This is defined as a transcript having an Adult-Child (AC) context.
% It is expected that a key reason why a system of this type has not been implemented on SMPs, or other platforms that allow for peer-to-peer communication that children commonly go on, is due to the lack of robustness that a system would have in determining OG specifically, and not similar language interactions such as 'Sexting' which is often described as happening between two adults.
While this information can be obtained within the sign-up process by using the date of birth, in a real-world context both the attacker and/or the child can lie about their true age to bypass any mitigation method that is dependent on this information with ease. However, by implementing a mitigation that focuses on language usage to determine age there is difficulty in an attacker mitigating this. Nevertheless, it should be noted that due to the variety of methods used by attackers, attackers can simulate being a child within their attack \citep{oconnell2003} and the language used is likely to reflect this. While this method would be out of the scope of this investigation, it is suggested that implementation to observe `language context differences' to identify if this method is being used, i.e. an actor context changing to an adult when the sexual stage \citep{oconnell2003}, could be used to identify this mitigation method. Therefore, it is theorised that the development of a method that can determine the context of a given interaction between two actors would provide a key piece of evidence in determining the complex concept of OG. In determining this context each of the two actors can be determined to be one of the following categories;
\begin{itemize}
    \item Significantly an Adult (A) 
    \item Significantly a Child (C)
    \item Not Significantly an Adult or Child (NS)
\end{itemize}
%The 'significance' element within this determination refers to the dynamic threshold in which an Adult or a Child categorisation shall be made under the Context Determination process. This will provide an adjustable confidence level that can be altered to observe the optimum confidence required for a specific social media platform use case. An Actor Significance P value threshold (AST) is used which refers to the probability that a given determination, as well as an equally likely/less likely determination, would occur by random chance. Differing AST values will be used within the investigation to observe their effects on the frequency of false positives and false negatives. We expect to see that with the less significant, higher AST values, a smaller or equal false negative frequency but a greater or equal false positive frequency will be obtained. Therefore with the more significant, lower, AST values we expect to see a greater or equal false negative frequency and a smaller or equal false positive frequency. It is likely in the context of SMPs that a smaller false positive frequency of OG determination is preferable, however it should be noted that in the context of this investigation the Adult/Child context insight is being determined - with the expectation for this insight to be combined with other insights to make an overall determination of OG. Therefore within this it is likely to be preferable to have a lower false negative frequency to allow for other insights to disprove a transcript as OG.

\begin{figure*}
    \centering
    \includegraphics[width=1\linewidth]{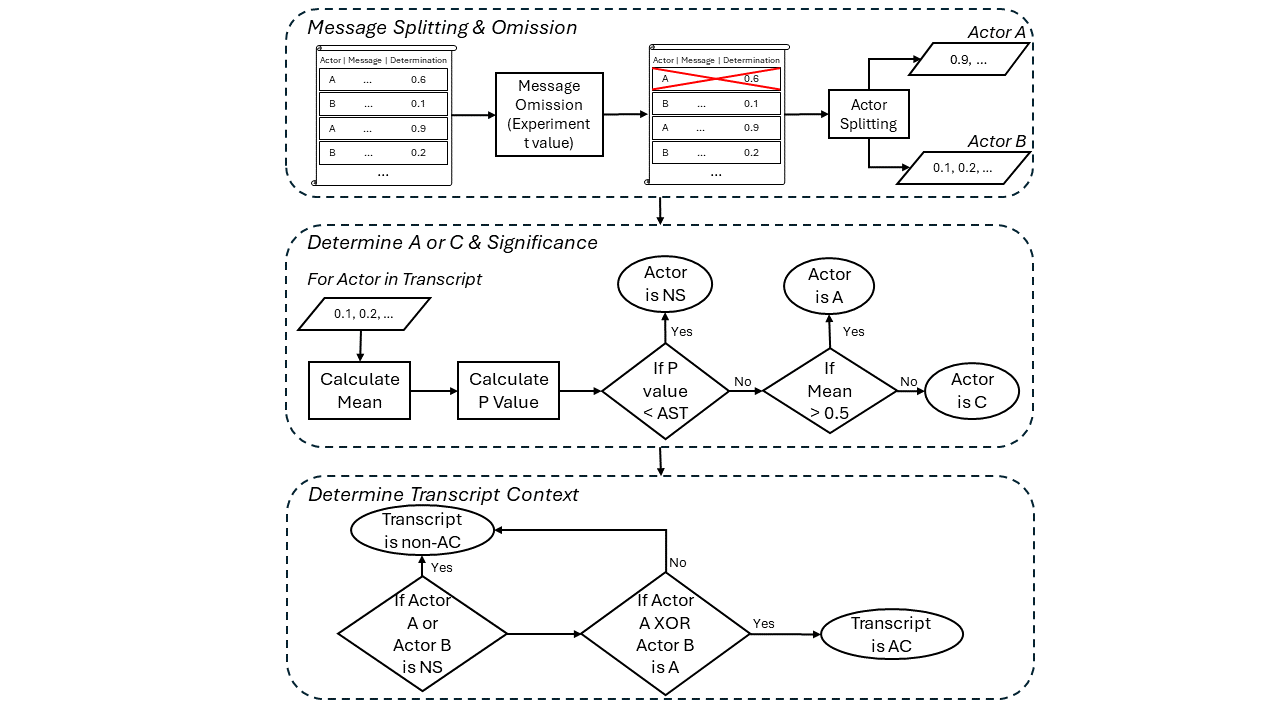}
    \caption{Context determination process with AST and $t$ value application.} 
    \label{fig:ContextDeter_Methodology}
\end{figure*}

The `significance' element in this determination refers to the dynamic threshold used in the Context Determination process to classify messages as Adult or Child. This threshold allows for adjustable confidence levels, which can be tailored to observe the optimal confidence needed for specific SMP use cases. An Actor Significance P value threshold (AST) is employed, representing the probability that a given determination, along with an equally or less likely determination, could occur by random chance. Different AST values will be tested to evaluate their impact on the occurrence of false positives and false negatives. It is anticipated that higher AST values, indicating less significance, will result in a lower or equal frequency of false negatives but a higher or equal frequency of false positives. Conversely, lower AST values, indicating greater significance, are expected to lead to a higher or equal frequency of false negatives and a lower or equal frequency of false positives.
In the context of SMPs, minimising the false positive frequency in OG determinations is typically preferable. However, it is important to note that in this investigation, the Adult/Child context insight is part of a broader determination of OG, expected to be combined with other insights. Therefore, a lower false negative frequency is likely preferable to allow other insights to refute a transcript as indicative of OG. These individual actor determinations can be mapped to an overall transcript context based upon the rule of combining the contexts unless one or both are deemed `Non Significant', in which case the full context determination is deemed Non Significant.
The methodology that shall be followed when using this approach is described within Figure \ref{fig:ContextDeter_Methodology} which demonstrates the three main processes in determining a given transcript's context. These are Message Splitting and Omission, in which each message is attributed to the actor that sent it while also omitting messages that are not significant enough based upon the $t$ value; Determining if each actor is an Adult or Child and the significance of this, where the AST experiment value is used to determine the threshold of significance; and Determine Transcript Context, in which the determinations from both actors are brought together to give an overall transcript context.
%This investigation focuses exclusively on peer-to-peer communications, where transcripts involve interactions between two unique actors. While there is a growing trend in the use of 'Group chat' features on SMPs, involving communication among more than two unique actors, the relevance of these 'Group Chats' to OG remains uncertain. Nonetheless, insights gained from this study could potentially be applied to this scenario in future research if deemed relevant.

% \begin{table}
%     \centering
%     \begin{tabular}{|c|c|c|c|} \hline 
%          &  \multicolumn{3}{|c|}{Actor 1}\\ \hline 
%          Actor 2&  A&  C& NS\\ \hline 
%          A&  Adult-Adult (AA) Context&  Adult-Child (AC) Context& Not Significant (NS) Context\\ \hline 
%          C&   Adult-Child (AC) Context&   Child-Child (CC) Context& Not Significant (NS) Context\\ \hline 
%  NS&  Not Significant (NS) Context& Not Significant (NS) Context&Not Significant (NS) Context\\ \hline
%     \end{tabular}
%     \caption{Transcript Context Outcomes}
%     \label{tab:fullCon}
% \end{table}

This investigation focuses exclusively on peer-to-peer communications, where transcripts involve interactions between two unique actors. While this approach raises some privacy concerns of the individuals using the platform to some degree, and with these interactions being of a sensitive nature users of all types would likely reject a system of this nature due to privacy concerns. Therefore the motivation of this work is intended to observe if the involvement of human reviewers, as seen with harmful content moderation, could be reduced in a potential implementation. By being able to reliably obtain the insight that an adult and a child are conversing, it is suggested that when this is applied alongside sexual language detection then a complex OG determination can be made which is likely to be robust against other messaging that is similar in language to OG e.g. `Sexting', defined as the consensual sending of sexual messages/media. As well as the proposed approach using AST value to determine the degree of confidence that is held for an Actor being an Adult or a Child, the same ideology can be applied to the individual messages that make up this determination. By using a `Message Threshold Value' messages that are of low confidence in determining an Adult or a Child can be omitted from consideration. This provides additional contextualisation to the problem of OG as not every message that is sent between two actors could be used to determine their age. 
A key point to note with the Message Threshold Value is that the frequency of low-confidence messages linked to an actor does not affect the overall actor determination process as these are simply omitted from consideration. This was an intentional choice as it is to be expected that both Adult and Child actors will send messages that do not significantly link to being either an Adult or a Child. Table \ref{tab:confidenceTransEg} shows an example transcript within the dataset with both low-confidence and high-confidence messages. As shown the first two messages display high Adult/Child confidence determinations i.e. it can easily be determined which actor is an adult/child, whereas, with the other two messages, there is a lack of confidence as it is likely that both an adult or a child could send the message. 
The `Prediction' values shown within Table \ref{tab:confidenceTransEg} were made from the BERT model.

%\begin{figure}
    %\centering
    %\includegraphics[width=1\linewidth]%{confidenceExample_Pan12-10.png}
    %\caption{Transcript Confidence Example within PAN12 %(dataframe format)}
%    \label{fig:confidenceTransEg}
%\end{figure}

\begin{table*}[htb!]
\scriptsize
  \centering
  \caption{Transcript confidence example within PAN12 (data frame format).}
  \label{tab:confidenceTransEg}
  {\begin{tabular}{l|l|l|l|l}
  \hline
 &\textbf{Authors}  & \textbf{OG} & \textbf{Message} & \textbf{Prediction} \\ \hline\hline
5 & 1 & 1 & im old 49 here old enough to be your daddy & 0.99998\\
6 & 0 & 0 & um my dad's older & 0.01687\\
19 & 1 & 1 & r u there & 0.38972 \\
20 & 1 & 1 & u have a good day & 0.32781 \\

\hline
  \end{tabular}}
\end{table*}

Message Threshold Values, $t$, from 0.2-0.45 shall be analysed within this investigation, this value will be applied to determine if a given message is to be considered within the Actor context determination. With a determination of `0.5' being the lowest confidence message determination value that could be given. A message will be omitted from consideration if Equation \ref{eq:1} is true, where $n$ is the determination value of the given message.
\begin{equation}
    (0.5 + t) > n > (0.5 - t)
    \label{eq:1}
\end{equation}

Only four set $t$ values shall be used within this investigation. However, it is expected that the optimum $t$ value could be obtained from regression if these $t$ values are found to have a significant effect on the overall Context Determination. The purpose of the Full Transcript Context Determination is to evaluate the validity of using an insight such as Adult/Child context interaction to determine OG, with future insights. All messages within a transcript will be utilised in this determination, therefore a message towards the beginning of the communication will have the same weighting in this analysis as one towards the end. Where $t$ is the message threshold used to be included within the determination (only applicable to BERT and RoBERTa). 
As depicted in Figure \ref{fig:ContextDeter_Methodology} Actor A and Actor B reference two unique actors within a peer-to-peer communication, Actors are then determined to either be `A', significantly an Adult; `C' significantly a Child; or `NS' not significantly either. Based upon that a check is made to see if there is an Adult Child combination using the following logic as shown within the `Determine Transcript Context' process described in Figure \ref{fig:ContextDeter_Methodology}: (Actor A is Adult XOR Actor B is Adult) AND (Actor A is Child XOR Actor B is Child).

% \begin{algorithmic}
    
% \end{algorithmic}

\section{Experimental Methodology}

This section shall give an overview of the implementation choices made as part of this investigation, as well as highlighting challenges faced within the preparation of the datasets.

\subsection{Datasets} 

Due to the ethical and legal considerations that must be made within obtaining OG transcripts, there is a scarcity of data that can be used.
The majority of literature investigating OG either uses the PAN12 dataset or a version of the Perverted Justice dataset including transcripts from 2006 up to 2016. However, most investigations only use a limited portion of the available data e.g. in the investigation by \citet{kontostathis2010}. This paper uses all available transcripts from the PJ dataset \footnote{perverted-justice.com} (at the time of publication) which have followed a semi-automated labelling process by the research team. % This up to date dataset is available at: ...
The transcripts were published using different HTML formatting over time the use of HTML parsing allowed for the majority of transcripts to be automatically labelled, as predators were identified using different styling or the predator's username was in the web page title. Nevertheless, some transcripts required manual labelling from the research team.
There have been investigations, of both a psychological and technical nature, using other datasets that have not been made publicly available due to these aforementioned considerations \citep{kloess_2015_a,cheong_2015_detecting,chiang_2019_deceptive}. Therefore within this investigation, the two datasets shall be used - from PJ and the PAN12 dataset \footnote{https://pan.webis.de/clef12/pan12-web/sexual-predator-identification.html}\citep{inches2012}. 
In contrast to existing OG detection literature, this study utilises the `full version' of the PJ dataset, which includes all available data up to the present, unlike previous studies that could only access datasets available at the time of their publication. It is important to note that 80\% of an earlier iteration of the PJ dataset (spanning 2006-2012) is included in the PAN12 dataset \citep{inches2012}, alongside transcripts from three other sources.
The PJ dataset distinguishes itself by labelling each transcript with a groomer and a victim for every message, facilitating the classification of each message as either Adult or Child based on these designations. However, in the PAN12 dataset, such labelling is absent for `negative OG' transcripts, which hinders the ability to contextualise these interactions. Nevertheless, the same methodology is employed in PJ can be applied to `positive OG' transcripts within PAN12.
Since the PJ dataset consists of .html webpages from the host website, a preprocessing and labelling step is necessary to prepare them for analysis. Each webpage contains a transcript with a dynamic preamble, necessitating HTML parsing to extract messages and essential information for identifying the attacker and victim in each transcript. This process involves examining the file name, typically indicating the attacker, although, in some instances, attackers may use multiple usernames, rendering this approach unreliable. Each transcript must undergo careful analysis to determine its formatting requirements, ensuring the correct processing steps are applied. Algorithm \ref{Alg:PJ_process} outlines this 

\begin{algorithm}
\begin{algorithmic}[1]
    \State $knownAttackers \gets AttackersFile$
    \For{$htmlPage$}
        \State $allText \gets htmlPage.read$
        \State $attackerTitle \gets htmlPage.replace(fileExtension)$
        \If{$blueBold \in allText$}
            \State $FormattedProcessing$
        \ElsIf{$attackerTitle \in allText \mid attackerTitle \in knownAttackers$}
            \State $NameProcessing$
        \Else
            \State $AttackerEntry$
        \EndIf
    \EndFor
    \caption{`peverted-justice.com' (PJ) webpage processing.}\label{Alg:PJ_process}
\end{algorithmic}
\end{algorithm}

As shown by Algorithm \ref{Alg:PJ_process}, a known attacker's file is used, this allows for the semi-automated labelling of messages which ensures that each message does not have to be labelled, and instead, the attacker name is searched for amongst the message lines. These attackers are entered into this file when the `attacker entry' process, as shown in Algorithm \ref{Alg:PJ_process}, is run. 
Following this each transcript is then checked for the `blueBold' formatting, which purely looks at this to determine attacker message lines and if this is not present name processing is used which utilises the `
`knownAttackers' and HTML file titles to attempt to label the attacker lines. 
To ensure that there are no errors within this a check on the number of messages for both labels is considered, and if there is a significant difference (more than 70\% of one label) between the labelling frequency within a transcript this is then manually checked and `attacker entry' process is then used to make adjustments.

\begin{figure}
    \centering
    \includegraphics[width=1\linewidth]{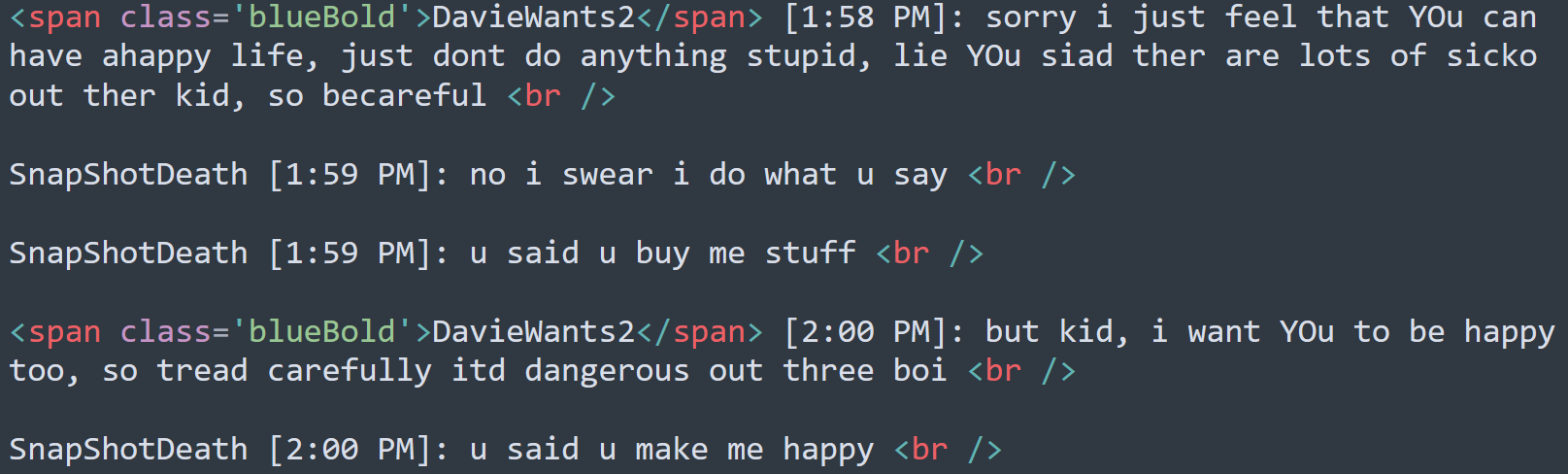}
    \caption{Example of a transcript for Formatted Processing (user `DavieWants2' from PJ).}
    \label{fig:PJformattedExample}
\end{figure}

\begin{figure}
    \centering
    \includegraphics[width=1\linewidth]{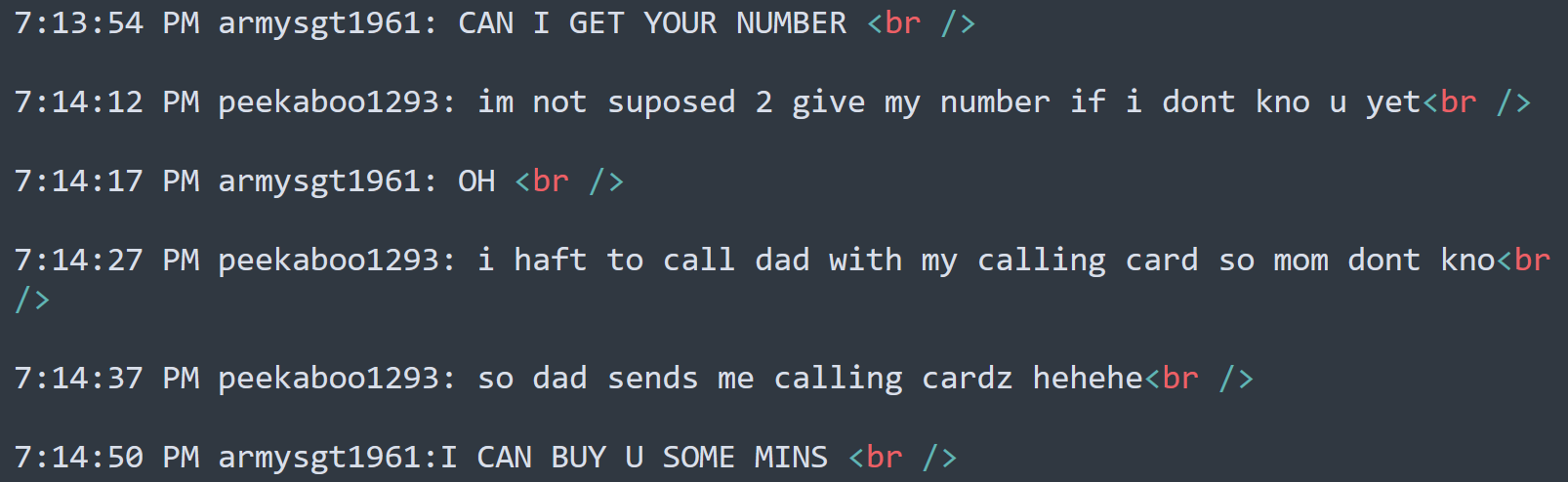}
    \caption{Example of a transcript for Name Processing (user `ArmySgt1961' from PJ).}
    \label{fig:PJnameExample}
\end{figure}

\begin{figure}
    \centering
    \includegraphics[width=1\linewidth]{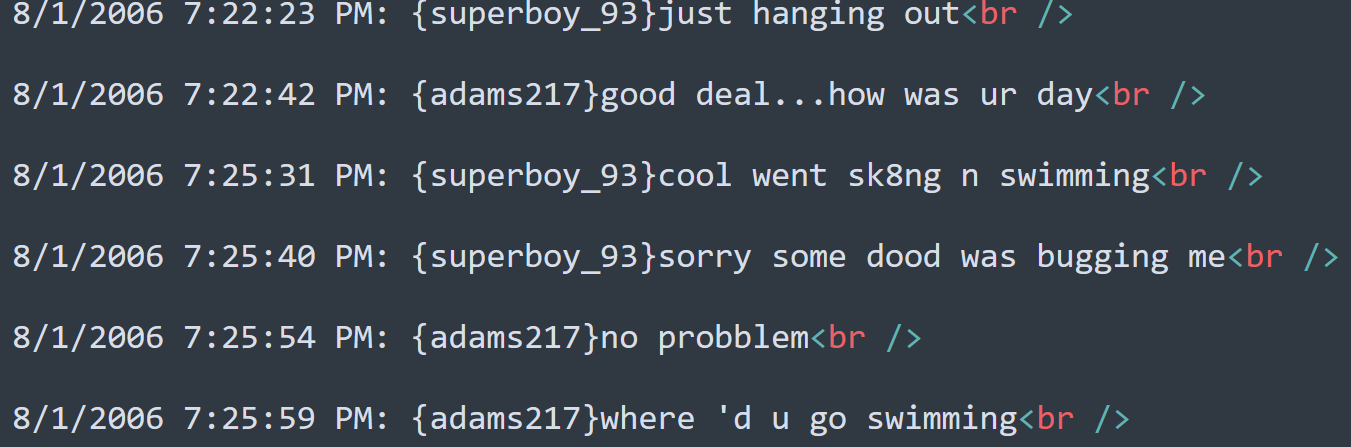}
    \caption{Example of a transcript for Attacker Entry (user `Adamou217' from PJ).}
    \label{fig:PJentryExample}
\end{figure}

As shown by the webpage example in Figure \ref{fig:PJformattedExample} there is `blueBold' formatting on the message lines that have been sent by the attacker, but this formatting is not present for the honeypot message lines. As shown by Figure \ref{fig:PJnameExample} there is an exact match on the message line of the attacker's username with the messages they have sent, however as seen within Figure \ref{fig:PJentryExample} a different username is being used by the attacker `adams217' compared to the title of the webpage `Adamou217' therefore attacker entry is needed to include `adams217' to the list of attackers. In addition, different formatting options were used on these webpages there were inconsistent prefixes to the message lines on these webpages. For example, some message lines include the time, some include `AM/PM', some include different brackets for formatting, and some include comments on specific messages that were not included within the communication. As seen across Figures \ref{fig:PJformattedExample}, \ref{fig:PJnameExample}, and \ref{fig:PJentryExample} there are differences in this time formatting with some including the full date, and some including the time to the resolution of seconds.

%% TODO - Is below needed?
% To ensure that only the message was included within the labelling, the following Regular expressions were run (in chronological order).
% \begin{enumerate}
%     \item 
% \end{enumerate}

% date_time_brac = "[\(\[][0-9][0-9]?[/-][0-9][0-9]?[/-][0-9][0-9][0-9]?[0-9]?  ?[0-9][0-9]?:[0-9][0-9]?:?[0-9]?[0-9]? [aApP][mM][\)\]]"
% time_brac = "[\[\(][0-9][0-9]?:[0-9][0-9]:?[0-9]?[0-9]?  ?[aApP][mM][\)\]]"
% time = "[0-9][0-9]?:[0-9][0-9] [aApP][mM]"
% comment = "<span class='code_c'>.+</span>"
% comment2 = "\(.+\)"
% comment3 = "\*.+\*"

\subsection{Preprocessing Methods}
Similar to the investigation by \citet{villatorotello2012} preprocessing methods will not be used in our approach, beyond the exclusion of transcripts shorter than 10 messages in length for the PAN12 dataset and the exclusion of email transcripts within the PJ dataset used in this investigation. It should be noted that, while this may improve the accuracy of determinations, this is unlikely to be transferable/robust to a new age real-world context due to the move away from the use of emoticons formed of punctuation marks, e.g. `(:-*)' for kiss \citep{villatorotello2012}, that are likely being used within these OG determination methods.

\subsection{Models} 
Naive Bayes (NB) and Support Vector Machine (SVM) have commonly been seen implemented as solutions to the binary classification of OG. In terms of the technical configurations used within this investigation, for NB the `sklearn.naive\_bayes' module was used \citep{scikit-learn} alongside the feature extraction module as part of this module. For the SVM model, this same module was used alongside the `pipeline' with a `countVectoriser' approach \citep{scikit-learn}. For both the transformer-based models (BERT and RoBERTa) `tensorflow' was utilised \citep{tensorflow}. For RoBERTa the `RoBERTa-base' tokeniser was used \citep{RoBERTa_tokeniser}, whereas for BERT the BERT tokeniser within tensorflow was used \citep{tensorflow}. 

\section{Results and Discussion}

The results within this investigation are split into two parts Message Level Analysis and Context Determination. The purpose of these two sections is to first, within Message Level Analysis, establish the baseline accuracy of models in both an inter-set and cross-set context, as well as to observe the impact of $t$ values. Then the second section, Context Determination, allows for a comparison to be made against this established baseline to put this in the real-world context of OG detection, as well as observing the impact that AST values have on this.  

\subsection{Message Level Analysis Results}
Across the PJ sample test set that was trained on the remaining 80\% of the PJ dataset, the MLA results are shown within Table \ref{tab:MLApjSample}. It should be remembered that despite the size of the PAN12 dataset in comparison to PJ, all messages within negative OG transcripts are excluded from MLA. Therefore in comparing the performances across both test sets this needs to be taken into account, as well as the increased training size for the models. Within Table \ref{tab:MLApjSample} `TP' refers to True Positive, `IA' refers to Incorrect Adult Determination, `IC' refers to Incorrect Child determination, and `O' refers to Omissions. Note that the percentage values do not sum to 100\% as the Omissions percentage value refers to the total percentage of messages that were omitted from consideration whereas the other percentage values refer to the respective metric of the messages that were still included within the process.
A total of 219,168 messages were included within this MLA analysis over the PJ dataset.

\begin{table}[]
\centering
\caption{MLA results across PJ sample.}
\label{tab:MLApjSample}
{\begin{tabular}{@{}ccccc@{}}
\toprule
Model& TP & IA& IC&O\\
\midrule
SVM& 76.0\%& 13.5\%& 10.5\%&0\%\\ 
NB& 74.7\%& 12.0\%& 13.3\%&0\%\\
\midrule
BERT (t=0.2)& 82.2\%& 9.3\%& 8.5\%&9.2\%\\
BERT (t=0.3)& 83.7\%& 8.4\%& 7.7\%&15.3\%\\
BERT (t=0.4)&  85.6\%& 7.8\%& 6.6\%&22.7\%\\
BERT (t=0.45)& 87.3\%& 7.1\%& 5.6\%&30.0\%\\ 
\midrule
RoBERTa (t=0.2)&83.3\%& 7.7\%& 9.0\%&8.8\%\\ 
RoBERTa (t=0.3)& 84.9\%& 6.9\%& 8.2\%&14.7\%\\ 
RoBERTa (t=0.4)& 87.2\%& 6.0\%& 8.9\%&23.6\%\\ 
RoBERTa (t=0.45)& 88.8\%& 5.3\%& 5.9\%&30.4\%\\
\bottomrule
\end{tabular}}
\end{table}
 
Table \ref{tab:MLApan12} shows the MLA results from the PAN12 dataset on transcripts that contain more than 30 messages and contain messages deemed to be predatory (assumed to be Adult), 
A total of 702,419 messages were included within this MLA analysis over the PAN12 dataset.

\begin{table}[]
\centering
\caption{MLA results across PAN12 (positive OG transcripts).}
\label{tab:MLApan12}
{\begin{tabular}{@{}ccccc@{}}
\toprule
Model& TP& IA& IC&O\\ 
\midrule
SVM& 63.4 \%& 25.5 \%& 11.2 \%&0.0 \%\\ 
NB& 61.4 \%& 23.4 \%& 15.2 \%&0.0 \%\\ 
\midrule
BERT (t=0.2)&64.3 \%& 25.8 \%& 9.9 \%&7.3 \%\\ 
BERT (t=0.3)&64.8 \%& 26.0 \%& 9.2 \%&11.1 \%\\ 
BERT (t=0.4)& 65.6 \%& 26.3 \%& 8.1 \%&16.3 \%\\ 
BERT (t=0.45)& 66.4 \%& 26.7 \%& 6.9 \%&20.8 \%\\
\midrule
RoBERTa (t=0.2)&65.1 \%& 26.6 \%& 8.3 \%&7.5 \%\\ 
RoBERTa (t=0.3)& 65.7 \%& 26.8 \%& 7.5 \%&12.2 \%\\ 
RoBERTa (t=0.4)& 66.5 \%& 26.9 \%& 6.6 \%&20.2 \%\\ 
RoBERTa (t=0.45)& 67.2 \%& 27.0 \%& 5.8 \%&26.3 \%\\
\bottomrule
\end{tabular}}
\end{table}

The results from the MLA on PAN12 as shown within Table \ref{tab:MLApan12} initially suggest that these models are not robust in determining Adult/Child across datasets, weighted significantly to incorrect Adult determinations. Following these results, the weightings between both labels in the PAN12 dataset were analysed and it was found that 68\% of the messages were labelled as predatory. This value seems unlikely to be a fair reflection of the transcripts in terms of Adult/Child interaction due to how great the difference is when considering the vast size of the dataset. 

% It should be noted that there is likely to be more polarisation of these test sets in OG positive than OG negative, this in turn could be used as another metric/approach within each transcript however further analysis would be needed. ???

When comparing the MLA results across PAN12 and PJ there is a significant difference in results which leads to the initial conclusion that cross-dataset determinations lack robustness. Despite this difference, it is theorised that when combining these determinations as part of full context determination may allow for an increase in overall accuracy. An interesting observation to be made is the True Positive Accuracy of SVM without any omissions. It is possible that if there were no omissions (i.e. a $t$ value of 0 is used) within the transformer-based models, SVM in this instance may outperform these models. Another important thing to note is the skew towards Adult determinations, as seen with the Incorrect Adult determination metric, within the PAN12 dataset. The mean difference between the sum of Incorrect Adult and Incorrect Child determinations within the PJ sample set was -0.02\% whereas the mean difference for PAN12 between Incorrect Adult and Incorrect Child was +17.4\%.
% PJ IA - 8.4
% PJ IC - 8.42
% PAN12 IA - 26.1
% PAN12 IC - 8.7
It is suggested that to counteract the `skew' that is apparent when conducting this cross-dataset analysis, dynamic $t$ values could be used for both the Adult and Child thresholds for omission. In this case, an inflated $t$ value for the Adult determinations (determinations more than 0.5) could be used with the expectation to reduce this `skew'. This is based on the assumption that the Incorrect Adult Determinations are closer in value to the omission threshold than the True Positive adult determinations. A further investigation is required to validate if this $t$ value adjustment method, and assumptions about the Incorrect Adult Determinations, have validity.

\subsection{Context Determination Results}

\subsubsection{PJ Results}
The below results are for PJ 20\% with a total of 152 transcripts which all have a context of `AC', therefore only the True Positive and False Negative metrics are used. Table \ref{tab:PJ20_Full05} gives an overview of the results obtained. 
\begin{table}[]
\centering
\caption{PJ sample full transcript context determination with an AST value of 0.05.}
\label{tab:PJ20_Full05}
{\begin{tabular}{@{} cccc@{} }
\toprule
Model&  TP&  FN& F1 \\
\midrule
SVM&  118&  35& 0.871\\
NB&  116&  37& 0.862\\
\midrule
BERT (t=0.2)&  143&  10& 0.966\\
BERT (t=0.3)&  141&  12& 0.959\\
BERT (t=0.4)&  141&  12& 0.959\\
BERT (t=0.45)&  141&  12& 0.959\\
\midrule
RoBERTa (t=0.2)&  142&  11& 0.963\\
RoBERTa (t=0.3)&  143&  10& 0.966\\
RoBERTa (t=0.4)&  144&  9& 0.970\\
RoBERTa (t=0.45)& 145& 8&0.973\\
\bottomrule
\end{tabular}}
\end{table}

\begin{table}[]
\centering
\caption{PJ sample full transcript context determination with an AST value of 0.01.}
{\begin{tabular}{@{} cccc@{} }
\toprule
Model&  TP&  FN& F1\\
\midrule
SVM&  114&  39& 0.854\\
NB&  112&  41& 0.845\\
\midrule
BERT (t=0.2)&  140&  13& 0.956\\
BERT (t=0.3)&  140&  13& 0.956\\
BERT (t=0.4)&  140&  13& 0.956\\
BERT (t=0.45)&  141&  12& 0.959\\
\midrule
RoBERTa (t=0.2)&  142& 11& 0.963\\
RoBERTa (t=0.3)&  142& 11& 0.963\\
RoBERTa (t=0.4)&  143& 10& 0.966\\
RoBERTa (t=0.45)& 143& 10& 0.966\\
\bottomrule
\end{tabular}}
\label{tab:PJ20_Full01}
\end{table}

\begin{table}[]
\centering
\caption{PJ sample full transcript context determination with an AST value of 0.001.}
\label{tab:PJ20_Full001}
{\begin{tabular}{@{} cccc@{} }
\toprule
Model&  TP&  FN& F1\\
\midrule
SVM&  106&  47& 0.819\\
NB&  107&  46& 0.823\\
\midrule
BERT (t=0.2)&  139&  14& 0.952\\
BERT (t=0.3)&  139&  14& 0.952\\
BERT (t=0.4)&  138&  15& 0.948\\
BERT (t=0.45)&  138&  15& 0.948\\
\midrule
RoBERTa (t=0.2)&  139&  14& 0.952\\
RoBERTa (t=0.3)&  139&  14& 0.952\\
RoBERTa (t=0.4)&  139&  14& 0.952\\
RoBERTa (t=0.45)& 140& 13&0.956\\
\bottomrule
\end{tabular}}
\end{table}

% \begin{figure*}
%     \centering
%     \includegraphics[width=1\linewidth]{PJ20_full_FN.png}
%     \caption{PJ Sample Full Context Determination FN Frequencies}
%     \label{fig:PJ_FULLfn}
% \end{figure*}
\begin{figure*}
    \centering
    \begin{tikzpicture}
        \begin{axis}[
        width = \textwidth*0.8,
        major x tick style = transparent,
        ybar,
        bar width=8pt,
        ymajorgrids=true,
        ylabel = {False Negative Frequency},
        symbolic x coords = {NB, SVM, BERT20, BERT45, RoBERTa20, RoBERTa45},
        label style = {font=\tiny},
        tick label style = {font=\tiny},
        xtick=data,
        scaled y ticks = false
        ]
        
        \addplot[style={bblue,fill=bblue,mark=none}]
        coordinates{(NB,37) (SVM,35) (BERT20,10) (BERT45,12) (RoBERTa20,11) (RoBERTa45,8)};
    
        \addplot[style={rred,fill=rred,mark=none}]
        coordinates{(NB,41) (SVM,39) (BERT20,13) (BERT45,12) (RoBERTa20,11) (RoBERTa45,10)};
    
        \addplot[style={ggreen,fill=ggreen,mark=none}]
        coordinates{(NB,46) (SVM,47) (BERT20,14) (BERT45,15) (RoBERTa20,14) (RoBERTa45,13)};
        
        \legend{AST = 0.05, AST = 0.01, AST = 0.001}
        \end{axis}
    \end{tikzpicture}
    \caption{PJ sample full context determination FN frequencies.}
    \label{fig:PJ_FULLfn}
\end{figure*}
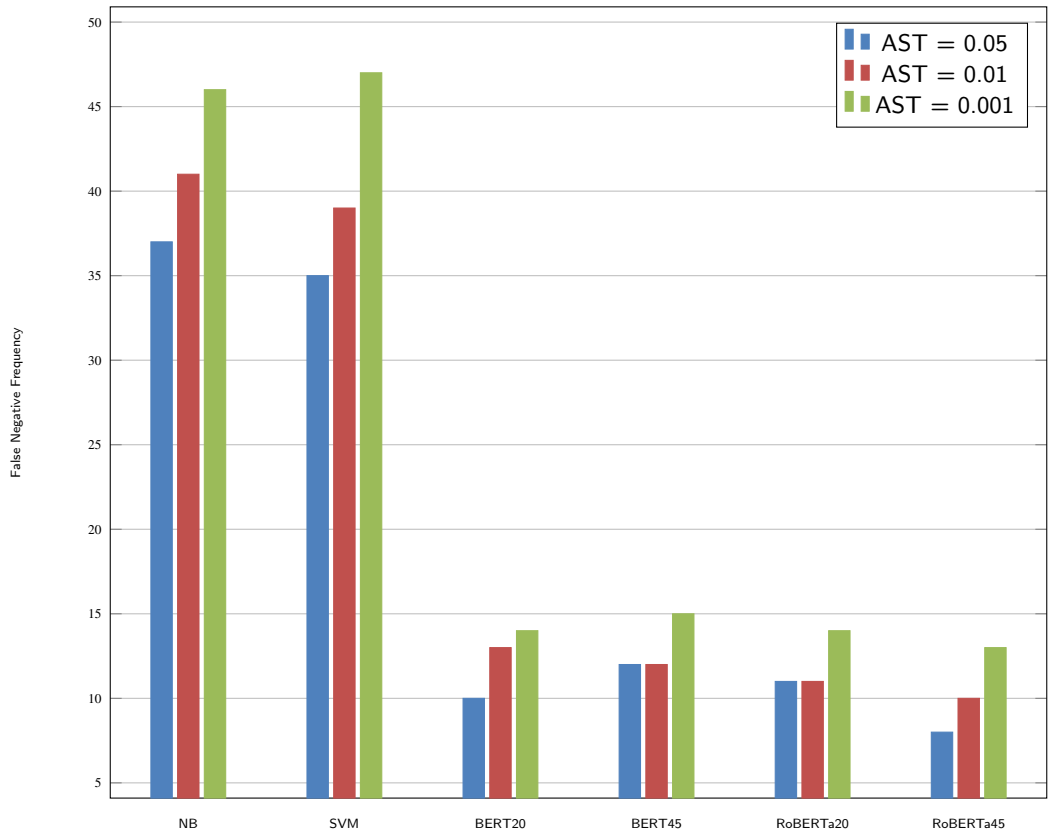
Figure \ref{fig:PJ_FULLfn} shows the False Negative frequencies amongst different models and message threshold values ($t$). These $t$ values are denoted in both Figures \ref{fig:PJ_FULLfn} and \ref{fig:PAN12_FNFP} within the model labels that have the suffix `-20', denoting a $t$ value of 0.2, and `-45', denoting a $t$ value of 0.45. As described the lower the Actor Significance Threshold (AST) value the reduction in False Negatives. It is anticipated that there will be a noticeable decrease in False Negatives when the message threshold value is increased for RoBERTa. Conversely, the opposite effect seems to occur when this approach is applied to BERT, as evidenced by the Actor Significance p-value of 0.05.
It is difficult to make conclusions on the application of these results without observing the False Positive rate of these models, which is unable to be done within the PJ sample as all transcripts are OG positive. This shall be investigated within the analysis of PAN12.

\subsubsection{PAN12 Results}
These results differ in terms of context when comparing these to the full context results for PJ where each transcript within PJ is classed as AC, this is not the case for PAN12 with the transcripts being any possible context. It should be noted that even OG negative transcripts could correctly be determined to be AC if the interaction is between an Adult and a Child but is not of a grooming nature. This is where further insight analysis, such as a sexual severity insight, would factor in to make a complete OG determination. Table \ref{tab:PAN12full05} shows the results from using an AST value of 0.05 on PAN12, emphasising reducing False Negatives (FN); Table \ref{tab:PAN12full01} shows the results from using an AST value of 0.01; and Table \ref{tab:PAN12full001} shows the results from using an AST value of 0.001, emphasising reducing False Positives (FP).

\begin{table}[]
\centering
\caption{PAN12 full transcript context determination with an AST value of 0.05.}
\label{tab:PAN12full05}
{\begin{tabular}{@{} cccccc@{} }
\toprule
Model&TP &FP &TN & FN &F1 \\
\midrule
SVM&717&27&  10309& 455 &0.748\\
NB&631&88&  10248& 541 &0.670\\
\midrule
BERT (t=0.2)& 1097& 27& 10309&75 &0.956\\
BERT (t=0.3)& 1098& 28& 10308& 74&0.956\\
BERT (t=0.4)& 1095& 28& 10308& 77&0.954\\
BERT (t=0.45)& 1092& 23& 10313& 80&0.955\\
\midrule
RoBERTa (t=0.2)& 1088& 26& 10310&84 &0.952\\
RoBERTa (t=0.3)& 1086& 26& 10310& 86&0.951\\
RoBERTa (t=0.4)& 1107& 27& 10309& 65&0.960\\
RoBERTa (t=0.45)& 1106& 26& 10310& 66&0.960\\
\bottomrule
\end{tabular}}
\end{table}

\begin{table}[]
\centering
\caption{PAN12 full transcript context determination with an AST value of 0.01.}
\label{tab:PAN12full01}
{\begin{tabular}{@{} cccccc@{} }
\toprule
Model &TP &FP &TN & FN &F1 \\
\midrule
SVM&548&19&  10317& 624&0.630\\
NB&475&52&  10284& 697&0.559\\
\midrule
BERT (t=0.2)& 1040& 14& 10322&132&0.934\\
BERT (t=0.3)& 1044& 14& 10322& 128&0.936\\
BERT (t=0.4)& 1042& 14& 10322& 130&0.935\\
BERT (t=0.45)& 1036& 13& 10323& 136&0.933\\
\midrule
RoBERTa (t=0.2)& 1039& 9& 10327&133&0.936\\
RoBERTa (t=0.3)& 1044& 10& 10326& 128&0.938\\
RoBERTa (t=0.4)& 1076& 9& 10327& 96&0.953\\
RoBERTa (t=0.45)& 1073& 15& 10321& 99&0.955\\
\bottomrule
\end{tabular}}
\end{table}

\begin{table}[]
\centering
\caption{PAN12 full transcript context determination with an AST value of 0.001.}
\label{tab:PAN12full001}
{\begin{tabular}{@{} cccccc@{} }
\toprule
Model &TP &FP &TN & FN &F1 \\
\midrule
SVM&388&18&  10318& 784&0.492\\
NB&344&40&  10296& 828&0.442\\
\midrule
BERT (t=0.2)& 939& 9& 10327&233&0.886\\
BERT (t=0.3)& 947& 10& 10326& 225&0.890\\
BERT (t=0.4)& 965& 11& 10325& 207&0.899\\
BERT (t=0.45)& 955& 10& 10326& 217&0.894\\
\midrule
RoBERTa (t=0.2)& 945& 5& 10331&227&0.891\\
RoBERTa (t=0.3)& 962& 6& 10330& 210&0.899\\
RoBERTa (t=0.4)& 1018& 7& 10329& 154&0.927\\
RoBERTa (t=0.45)& 1025& 10& 10326& 147&0.929\\
\bottomrule
\end{tabular}}
\end{table}

% \begin{figure*}
%     \centering
%     \includegraphics[width=1\linewidth]{PAN12_FNFP_edited.png}
%     \caption{PAN12 Full Context Determination FP \& FN Frequencies}
%     \label{fig:PAN12_FNFP}
% \end{figure*}

\begin{figure*}
    \centering

    \begin{tikzpicture}
        \begin{axis}[
        width = \textwidth,
        major x tick style = transparent,
        ybar,
        bar width=6pt,
        ymajorgrids=true,
        ylabel = {False Negative/Positive Frequency},
        symbolic x coords = {NB, SVM, BERT20, BERT45, RoBERTa20, RoBERTa45},
        label style = {font=\tiny},
        tick label style = {font=\tiny},
        xtick=data,
        scaled y ticks = false
        ]
        
        \addplot[style={ggreen,fill=ggreen,mark=none}]
        coordinates{(NB,88) (SVM,27) (BERT20,27) (BERT45,23) (RoBERTa20,26) (RoBERTa45,26)};
    
        \addplot[style={rred,fill=rred,mark=none}]
        coordinates{(NB,541) (SVM,455) (BERT20,75) (BERT45,80) (RoBERTa20,84) (RoBERTa45,66)};
        
        \addplot[style={bblue,fill=bblue,mark=none}]
        coordinates{(NB,52) (SVM,19) (BERT20,14) (BERT45,13) (RoBERTa20,9) (RoBERTa45,15)};
    
        \addplot[style={oorange,fill=oorange,mark=none}]
        coordinates{(NB,697) (SVM,624) (BERT20,132) (BERT45,136) (RoBERTa20,133) (RoBERTa45,99)};
    
        \addplot[style={ppurple,fill=ppurple,mark=none}]
        coordinates{(NB,40) (SVM,18) (BERT20,9) (BERT45,10) (RoBERTa20,5) (RoBERTa45,10)};
        
        \addplot[style={tteal,fill=tteal,mark=none}]
        coordinates{(NB,828) (SVM,784) (BERT20,233) (BERT45,217) (RoBERTa20,227) (RoBERTa45,147)};
        
        \legend{AST = 0.05 FP, AST = 0.05 FN, AST = 0.01 FP, AST = 0.01 FN, AST = 0.001 FP, AST = 0.001 FN}
        \end{axis}
    
    \end{tikzpicture}
    \caption{PAN12 full context determination FP \& FN frequencies.}
    \label{fig:PAN12_FNFP}
\end{figure*}
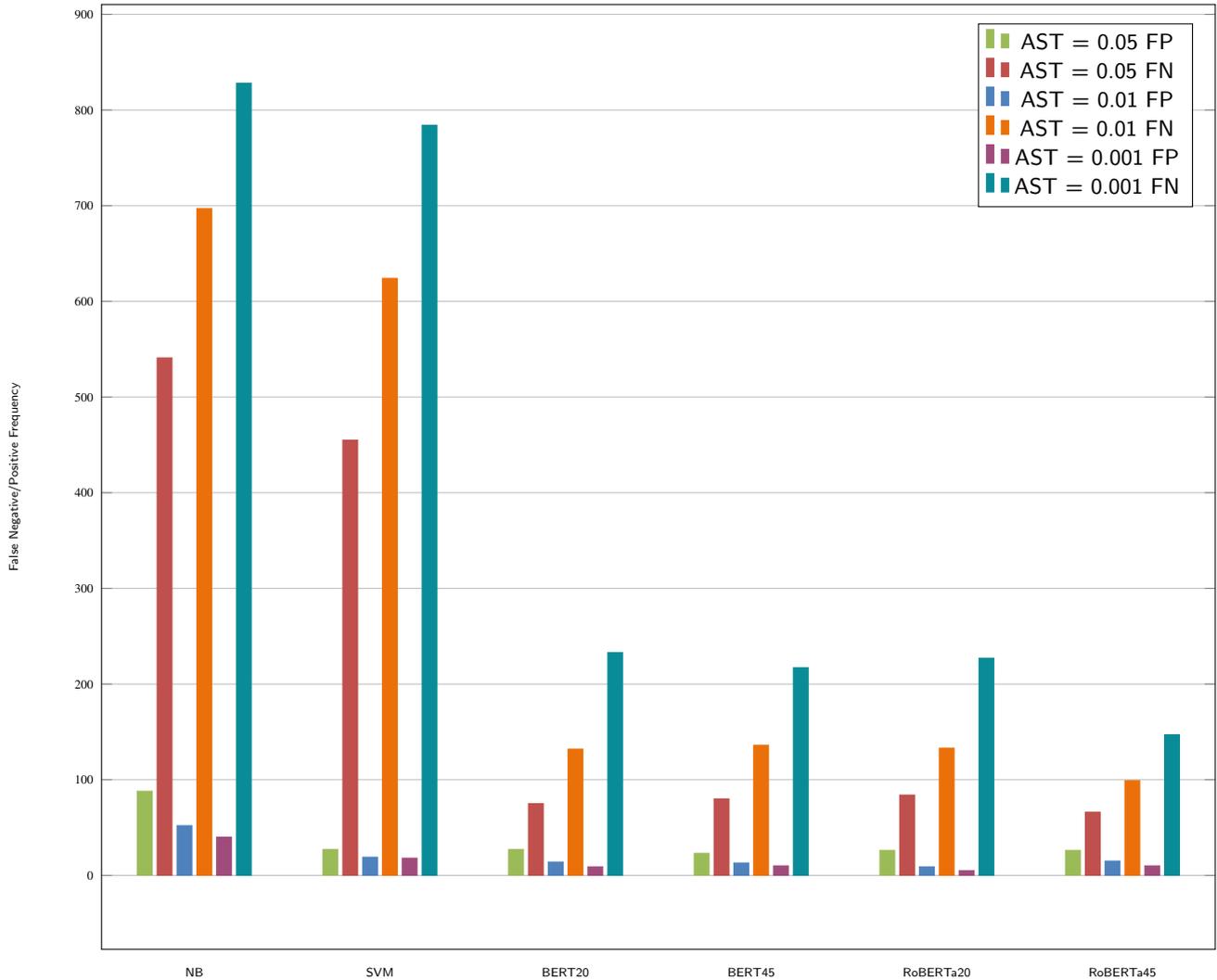

Figure \ref{fig:PAN12_FNFP} demonstrates the False Positives and False Negatives within the PAN12 dataset when using different models and $t$ values. From this it can be suggested that a greater $t$ value leads to greater accuracy of results, therefore greater weight should be made towards the key polarising tokens/messages that indicate either an adult or a child. 
In addition, it can be suggested that, as the False Negative frequencies are significantly greater than the False Positive frequencies amongst all models \& $t$ values, a less significant Actor Significance value could be implemented to observe the point of convergence between these two values to improve the accuracy of the context determination.

\subsection{Discussion}
Despite the poor results for the MLA on PAN12 the results from the full context determination of PAN12 suggest that this approach can be used to provide an OG determination with good accuracy. It is proposed that this approach in conjunction with other models can allow for the robust, complex determination of OG.
When considering the four models that were used in this investigation it is clear that the transformer-based models were better performing. This was clearest within the MLA results for the same dataset (over PJ sample) with a difference of up to 14.1\% in the rate of correct determinations (between NB and RoBERTa with $t=0.45$).
The results for MLA for robustness across the PAN12 dataset were somewhat similar across all models - in terms of correct determination percentage. However, it should be noted that with BERT and RoBERTa, due to the Message Threshold value, there are some omissions within the messages that could alone explain the percentage increase in accuracy as NB and SVM did not incorporate this process. In terms of the impact on the full context determination metric for the PJ sample, we can observe a difference in F1 score of up to 0.14 between NB and RoBERTa ($t=0.45$), when considering an AST value of 0.001; a difference in F1 score of up to 0.12 between NB and RoBERTa ($t=0.45$), when considering an AST value of 0.01; and a difference in F1 score of up to 0.11 between NB and RoBERTa ($t=0.45$), when considering an AST value of 0.05. Despite this reduction in the difference between F1 results across different AST values, the increase in AST value leads to better performance. This is also the case in increasing the $t$ value, most notably when using RoBERTa. It is suggested that increasing this AST value to observe the impact would be useful in finding the best solution, in the context of the PJ sample due to the lack of negative transcripts we are unable to confidently assume that this will increase the overall accuracy. However as the $t$ value does not directly affect this methodological concern in this experiment, this value can be adjusted closer to 0.5 to observe the impact this has on the overall accuracy.

In terms of the context determination for PAN12, in which robustness is being considered, we can see a significant difference in the F1 score between the non-transformer-based models and transformer-based models across the usage of all AST values. This is more prominent when using a smaller AST value. However, it can be observed that an F1 accuracy score of up to 0.96 from using the full PJ dataset in determining OG across PAN12, when using RoBERTa ($t=0.45$) and an AST of 0.05. This validates the claim that context determination is a robust approach that can be used in the determination of OG. Furthermore as shown by Figure \ref{fig:PAN12_FNFP} we can observe the low false positive frequencies in comparison to the false negative frequencies across all models. This suggests that an increase in AST value could lead to a further reduction in false negatives and, despite a probable increase in false positives,  should lead to an increase in the overall accuracy. This is in conjunction with using a greater $t$ value. It is suggested that an implementation of this approach would require other insights to provide further evidence/a bespoke approach for the SMP. Due to this proposed combination of insights it is suggested that each model/insight within this would allow for the question  `Does this outcome allow us to say that OG is not occurring?' to be answered. If this is the case then a reduction in the false negative frequency is preferable to ensure that true OG transcripts are dismissed and allow for false positive transcripts to be dismissed using one of the other insights in the proposed model.

In terms of the message threshold, it appears that for BERT this had minimal impact on the overall context determination results and the PAN12 MLA results. However in terms of the PJ sample experiment, there is a slight positive correlation between the $t$ value and the percentage of correct determinations, however, if we were to rely on this metric alone we are likely to not consider the fact that 30.0\% of the messages would be omitted (if using BERT, with a $t$ value of 0.45) therefore in terms of a realtime implementation this will require many more messages to make a determination and, in a real-world context, may cause the severe impacts of OG to occur before a determination can be made. This, however, is out of the scope of the full context transcript determination approach.

% Talk about RoBERTa and the impact of t

\subsubsection{Use Case Analysis}
In the use case of best performance, it is clear from the results obtained that using RoBERTa with an AST of 0.05 and a $t$ value of 0.45 is the best approach.
However, when considering the real-world implementation of this approach for SMPs this is dependent on if the results of this approach will be combined with another insight. If this approach were to be implemented alone then it would be preferable to seek for the lowest rate of False Positives to reduce the unnecessary breach of privacy/intervention from human reviewers that would be required as part of incident response. Therefore the results within this investigation suggest taking an approach with a lower $t$ value and higher AST value would provide the best solution to this (as demonstrated by RoBERTa $t=0.2$ with an AST of 0.001).
However in the case of this insight being used in conjunction with other insights, which is the intention for this approach, then subject to the results of these other insights and further analysis - it is preferable to have a lower False Negative score (which would require the use of the best-performing approach), if an `evidence to reject' approach is taken, or the greatest accuracy if a Fuzzy logic approach is applied to this complex OG determination approach.

\subsubsection{AST \& $t$ Value Regression for FP/FN Convergence}
Following the findings in this investigation in which RoBERTa is the best-performing model for context determination, it was observed that this was for the greatest AST value and $t$ values that were tested. Therefore it is possible that the optimum F1 score has not been reached. Therefore the AST and $t$ values will be increased to identify a point of convergence between False Positive Frequency and False Negative, which should give a better F1 score than 0.96.

Table \ref{tab:AST_t_rob} shows the F1 scores of this analysis and as shown the assumption made to increase the F1 score by increasing the AST value beyond what was previously tested was incorrect, however increasing the $t$ value to 0.47 some minimal increase in F1 score over using a $t$ value of 0.45.

\begin{table}[]
\centering
\caption{AST and $t$ value optimum for RoBERTa over PAN12.}
\label{tab:AST_t_rob}
{\begin{tabular}{@{} ccccc@{} }
\toprule
$t$ Value &AST 0.05 &AST 0.10 &AST 0.15 & AST 0.20\\
\midrule

0.45 & 0.960 & 0.960 & 0.958 & 0.958\\
0.47 & 0.961 & 0.960 & 0.959 & 0.959\\
0.49 & 0.954 & 0.956 & 0.956 & 0.951\\

\bottomrule
\end{tabular}}
\end{table}

\section{Conclusion and Future Work}

This investigation demonstrates that a context determination approach is significantly effective at determining AC interactions as seen within the same (F1 = 0.973) and different (F1 = 0.960) datasets. There is some uncertainty about the generalisability of this approach to non-OG AC contexts however analysis across different use cases can allow for this to be tested.
When considering the research questions on which this paper is based:
\begin{itemize}
    \item \textit{\textbf{Can a robust model be formed from taking a full transcript context determination approach?}}: While there is some difference between the F1 score obtained from taking the same dataset and a different dataset approach, this difference is minimal (-0.013) and unlikely to impact any real-world implementation of this approach in any meaningful way. This provides hope that this approach will be able to be used across many different social media platform contexts with a good level of effectiveness.
    \item \textit{\textbf{What impact do $t$ values and AST values have on the transcript OG determination accuracy?}}: Within this investigation, there was a clear trend that across the best performing model (RoBERTa) an increase in $t$ value led to an overall reduction in the False Negative frequency for the PJ sample experiments and the False Negative and False Positive frequencies across the PAN12 experiments. This suggests that it might be most effective to omit a high proportion of messages sent within this analysis and only rely on the most polarising messages. It should be noted in a real-world/modern-day context, transcripts may be shorter in length than the ones observed within the PJ and PAN12 datasets. As shown by the F score regression within Table \ref{tab:AST_t_rob} there seems to be a limit to the polarisation of these messages, as shown by the dip in F1 score when using a $t$ value of 0.49 (which equates to only considering messages between 1-0.99 and 0-0.01), with the best performing $t$ value of 0.47 (which equates to only considering messages between 1-0.97 and 0-0.03). It is expected that in a modern-day implementation, the `robustness gap' between PJ and PAN12 is smaller than modern-day OG attacks i.e. we expect to see that current attacks have greater linguistic differences to PJ and PAN12 than PAN12 to PJ. In terms of AST values, it can be seen that an increase in this variable leads to an increase in F1 score across both datasets. This appears to be the case due to the greater False Negative frequency in comparison to the False positive frequency that can be seen across the PAN12 experiments, and therefore while decreasing the AST value causes a fall in False Positives there is a smaller initial frequency (at AST = 0.05) compared to the False Negative difference (when comparing AST = 0.001 and AST = 0.05). If the percentage change of False Negatives/Positives as opposed to the frequency on the best-performing model (RoBERTa with t = 0.45) is considered, it can be observed that there is a 123\% increase in False Negative rate from using an AST of 0.001 as opposed to 0.05, and similarly, there is a 116\% increase in False Positive rate from using an AST of 0.05 as opposed to 0.001. This suggests that there is a similar overall effect of changing AST, which provides evidence that AST can be fine-tuned to a specific social media platform's use case, based on tolerances to False Positives/Negatives.
    \item \textit{\textbf{What model is recommended to be used in different use cases?}}: It is recommended that the transformer-based models are most appropriate for most use cases. The only consideration that should be made to this is computational power/prediction time that is required with these models may make these not fit for purpose and therefore the use of SVM or NB may be valid. However, it is suggested that while a decent F1 score of 0.871 was obtained with SVM across the PJ sample (when using AST = 0.05), it is clear that this approach lacks robustness as shown by the F1 value obtained across PAN12 for the same experiment (0.748), therefore it is expected that this approach would lead to an unmanageable number of false positives/negatives to make this feasible in a real-world setting.
\end{itemize}

% Contributions etc. - I don't think these need to be repeated here...
% \item Evaluating a novel 'context determination' approach that focuses on classifying actors and the interactions thereof, which is likely to allow generalisation to group chat settings
% \item Utilising message level analysis in an OG context using BERT and RoBERTa
% \item Conducting experiments on an updated version of the PJ dataset, with actor context labelling not previously used within literature
% \item Implementing the use of Actor Significance Thresholds and Message Significance Thresholds to more accurately fit the problem of OG
% \item Conducted cross-dataset experiments to evaluate the robustness of the message level analysis and context determination approaches

\subsection{Future Work}
Further work is required to provide a solution to OG detection, this predominately refers to the continued analysis of psychological literature to provide a robust solution that can detect the complex differences between the variety of attack methods that are used within OG.
These complexities refer to the assumed differences in language between Child Contact Sexual Offenders (CCSOs) and Fantasy Child Sexual Offenders (FSCOs) \citep{chiu2018}. It is assumed that CCSOs may not or briefly include sexual stages, as theorised by Ref. 13, and yet focus more so on the physical meet-up aspect. To contextualise this with the findings of this investigation, by combining context determination with a physical meet-up insight this could allow for the determination of a CCSO-type attack.
Conversely, it is expected that the physical meet-up stage will not be present within FSCO-type attacks, with an expectation that there is more reliance on the sexual and damage limitation stages \citep{oconnell2003}. Therefore combining the context determination results with an insight to determine sexual language/damage limitation, could allow for the determination of FSCO-type attacks. It should be noted that it is assumed that other insights can be used to improve the overall accuracy of OG determination, and less so provide useful attack method insights for incidence response purposes. These insights are likely to include the Romantic/Exclusivity stage and Risk Assessment stage \citep{oconnell2003}, however, the presence of these stages within current OG attacks is not known. To provide further validity to the claims of robustness it is suggested that the removal of punctuation and dated text speak may ensure that these findings are more generalisable to a real-world context, in determining an actor as an Adult or a Child. It is assumed that this experimental change will negatively impact the MLA, following the findings of Ref. 16. 
However, it is assumed that by utilising context determination and the adjustment of $t$ and AST values the impact this may have can be mitigated. A desirable outcome within a solution would be the ability for real-time detection. The analysis of sealed OG transcripts conducted by Ref. 23, found that the severe effects (e.g. sending of sexual media) of OG happen at around the 50-70\% point of the transcript. It should be noted that by taking a real-time approach these additional insights mentioned in this section would unlikely be established, and thus a greater reliance on the context determination method as described in this paper.
In addition, due to the increase in prevalence of group chat interactions on Social Media Platforms, the integration of Context Determination within this should be considered.

Further work should also analyse the presence and possible mitigations of the `skew effect' when comparing different dataset sources, as seen when comparing the results of MLA between PJ sample and PAN12.
 
% application with other insights (Oconnell2003)
% Language differences between CCSOs and other \cite{Chiu2018}
% Realtime - find source (broome) it is a lingustic source!
% Combination of insights
% Identifying methods to improve robust MLA accuracy by accounting for the "skew effect" across datasets?

\printcredits

\bibliographystyle{cas-model2-names}

\bibliography{cas-refs}


\end{document}